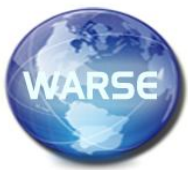

# A Smartphone-Based Skin Disease Classification Using MobileNet CNN

Jessica Velasco[1], Cherry Pascion[2], Jean Wilmar Alberio[3], Jonathan Apuang[4], John Stephen Cruz[5], Mark Angelo Gomez[6], Benjamin Jr. Molina[7], Lyndon Tuala[8], August Thio-ac[9], Romeo Jr. Jorda[10]

[1,2,3,4,5,6,7,8,9,10]Department of Electronics Engineering, Technological University of the Philippines, Manila 1000, Philippines

[1,2,9,10]Center for Engineering Design, Fabrication, and Innovation, College of Engineering, Technological University of the Philippines, Manila 1000, Philippines

**ABSTRACT**

The MobileNet model was used by applying transfer learning on the 7 skin diseases to create a skin disease classification system on Android application. The proponents gathered a total of 3,406 images and it is considered as imbalanced dataset because of the unequal number of images on its classes. Using different sampling method and preprocessing of input data was explored to further improved the accuracy of the MobileNet. Using under-sampling method and the default preprocessing of input data achieved an 84.28% accuracy. While, using imbalanced dataset and default preprocessing of input data achieved a 93.6% accuracy. Then, researchers explored oversampling the dataset and the model attained a 91.8% accuracy. Lastly, by using oversampling technique and data augmentation on preprocessing the input data provide a 94.4% accuracy and this model was deployed on the developed Android application.

## 1. INTRODUCTION

The Philippines being a tropical country has a tropical marine climate is divided into two main seasons: a rainy and a dry season. It has an average temperature of 27-33 degree Celsius as well as a relative humidity of 77-83%. These environmental factors and the coupled with age, occupation, and immune responsiveness contribute to the high prevalence of skin diseases among Filipinos.

Skin diseases are characterized as disorders that often begin inside the body or start from the skin, and outwardly show on the skin [1, 2]. Some of them are extremely uncommon, however, others are commonly occurring. They bring the person itch, and pain, as well as emotional and social impacts because of its visibility. All things considered, dermatologists guaranteed that a large portion of the skin diseases are manageable with legitimate medications, if they are decisively diagnosed [3]. Thus, an effective automatic skin disease detection design which can be used by the dermatologist to reduce their workload is highly anticipated.

In the recent years when big data started to boost like around 2011, the Big Data is frequently denoted by 5Vs: the wide variety of data types, the extreme volume of data, speed at which the data must be processed, the variability of the data, and the value of data [4, 5]. Big Data is a term used for any assembly of massive data sets whose complexity, as well as size, exceeds the capacity of conventional data processing applications [6, 7]. Clinical and epidemiology offers extraordinary research opportunities using Big Data to help create scientific advancements [8, 9]. Utilization of big data and image recognition technology along with the field of dermatology could yield remarkable aid to patients, dermatologist, as well as the research community [10, 11]. There are plenty of skin diseases which can be diagnosed by a dermatologist through ocular inspection. Utilization of artificial intelligence coupled with the underlying technology of deep learning for diagnosis is made reasonable by the fact that each of these conditions has a unique visual feature [12, 13]. Cases of skin diseases that are prevalent in the Philippines and potentially can be identified by image processing technologies are chicken pox, acne, eczema, pityriasis rosea, psoriasis, tinea corporis, and vitiligo.

The researchers in [14] implemented an Artificial Neural Network ANN-based single level system as well as a multi-model, multi-level system for eczema detection. Moreover, ANN was applied in [15] to detect certain circulatory diseases through the color of the fingernails. A similar implementation was applied in disease detection using tongue images [16]. A method of diagnosing Alzheimer's disease is by considering the retina images of the patients [17]. Meanwhile, a system was proposed using artificial neural network as well as digital image processing in detecting BCC disease [18]. The detection is based on special characteristics of basal cell carcinoma. The system will be capable to correctly identify the occurrence of carcinoma using the proper threshold values with percent reliability of 93.33%.

In [19], the proponents made use of the Sobel operator for the segmentation process. Three different skin diseases are selected: Psoriasis, Seborrheic Keratosis, and Pyoderma. Two feature sets were experimented on with one feature set having 86 color and texture features and the other having 4,182 color and texture features. The results are promising given that the average F-measure using 86 features was 88.67% while the average F-measure using 4,182 features was 84.81%.

In [20], the researchers used support vector machine (SVM) to classify melanoma skin cancer. They collected dermoscopy image database, segmented it using thresholding, collected unique characteristics, calculated total dermoscopy score and then classified it using SVM. The accuracy they got was 92.1%.

Lastly, in [21], the researchers concluded that deep learning algorithms are viable for diagnosing skin diseases. The aim of the study was to apply deep neural network algorithm in classification of four common skin diseases. The researchers developed the algorithm from GoogleNet Inception V3 package. They adjusted the final layer to add their own datasets using transfer learning. It had promising results having 86.54±3.63% accuracy using the first dataset and 85±4.649% for the second dataset.

We propose to visually infer the aforementioned skin diseases by harvesting images obtained from professional and publicly accessible websites, photo atlas of dermatology and taken manually and classify each image into the right skin disease category via transfer learning models.

In general, this study aims to design a skin disease classification system application in an Android phone that will classify different skin diseases using the highest performance pretrained convolutional neural networks model in the said field of dataset.

## 2. METHODOLOGY

### 2.1 Dataset

The dataset comes from a combination of public accessible dermatology repositories, color photo atlas of dermatology and taken manually. The images gathered from online public access dermatology repositories are validated by dermatologist. Data images gathered consist of acne, eczema, pityriasis rosea, psoriasis, tinea corporis, varicella(chickenpox) and vitiligo.

### 2.2 Data Preparation

The dataset contains sets of images in jpeg extension some taken manually corresponding to the same skin lesion but from multiple viewpoint or multiple set of images acquired on the same person. Figure 1 show some example of images for 7 skin diseases.

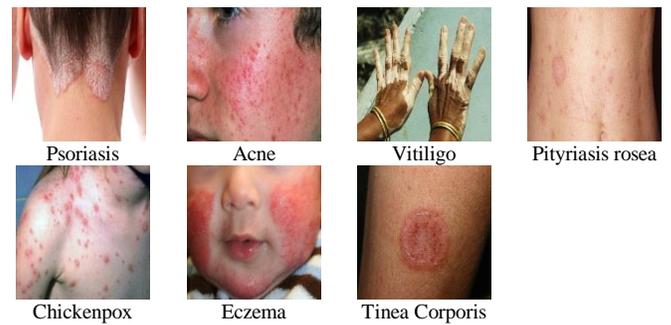

Figure 1: Sample Images of Dataset

### 2.3 Sample Selection

The dataset is partition to train and test data for each category of skin disease. The train data consist of 80% of the dataset while the test data is 20%. The validation data was gathered either in the train data or test data with the same number of images compared to test data as shown in Table 1.

Table 1: Dataset

| Skin disease | Number of Images | Train Data | Test Data | Validation Data |
|---|---|---|---|---|
| Acne | 805 | 644 | 161 | 161 |
| Eczema | 599 | 479 | 120 | 120 |
| Chickenpox | 314 | 251 | 63 | 63 |
| Pityriasis rosea | 347 | 278 | 69 | 69 |
| Psoriasis | 633 | 506 | 127 | 127 |
| Tinea Corporis | 385 | 308 | 77 | 77 |
| Vitiligo | 323 | 258 | 65 | 65 |
| **Total** | **3,406** | **2,724** | **682** | **682** |

### 2.4 Training Algorithm

Based on the Table 2, the proponent decided to use the pretrained CNN MobileNet model. The final classification layer of the model was removed and freeze the other layer and retrain the last layer with our dataset, fine tuning the parameters of the said model. Preprocessing the input images with a size of 224x224x3 pixels. To fit the model the following parameters was used: the learning rate is 0.0001, the activation is softmax, the loss is categorical crossentropy, the optimizer is Adam and the epoch is 30.

Table 2: Evaluation of MobileNet Model by its Weight size, Loading time and Accuracy

| Model | Weight size | Loading Time (seconds) | Accuracy (Percent) |
|---|---|---|---|
| MobileNet | 16.823MB | 4.838 | 84.28 |

## 2.5 Inference algorithm

After training the model, the model weight and architecture can be saved with .h5 file extension which is a Keras file. To properly deploy the CNN model to an Android application the .h5 file should be converted to a protobuff file with a file extension of .pb. Tensorflow is needed in this process where you freeze the graph producing a .pb file and a txt file containing the labels of the created model. To fully use the model for inference, a Tensorflow utility is used which is the optimize_for_inference function and with this the model can be successfully loaded to your Android application. Figure 2 shows the block diagram of necessary steps in deploying the model.

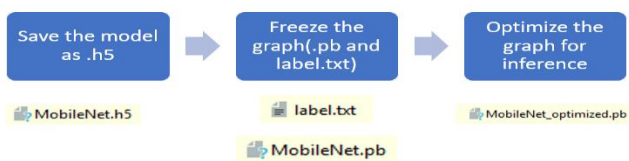

**Figure 2:** Deploying the Model

## 2.6 ANDROID APPLICATION

The development of an Android smartphone application that can identify skin diseases using deep learning through Convolutional Neural Network has three main process. These are capturing the skin lesion, deep learning analysis, and displaying the result, as shown in Figure 3 and the graphic user interface of the application is shown in Figure 4.

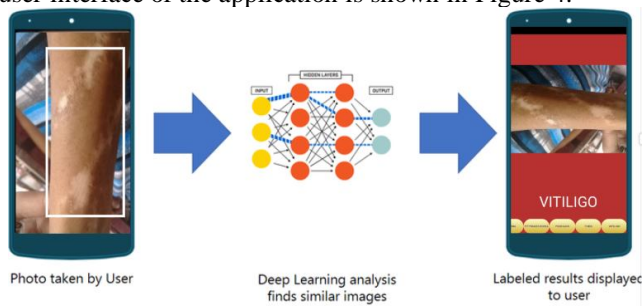

**Figure 3:** Overview diagram of the system

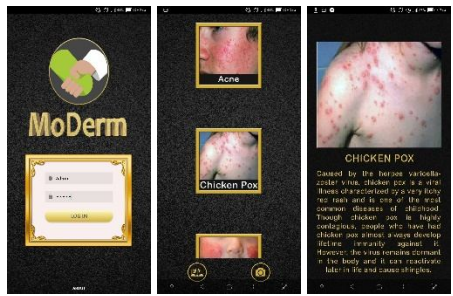

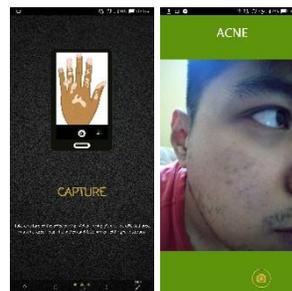

**Figure 4:** GUI of the Android application

## 3. RESULTS AND DISCUSSION

### 3.1 MobileNet versions

The MobileNet model was trained on using the imbalanced dataset and the default preprocessing of input data. Figure 5-6 show the confusion matrix with an accuracy of 93.6%. Table 3 shows how the train data and test data partition.

**Table 3:** Imbalanced dataset

| Skin disease | Number of Images | Train Data (80%) | Test Data (20%) | Validation Data |
|---|---|---|---|---|
| Acne | 805 | 644 | 161 | 161 |
| Eczema | 599 | 479 | 120 | 120 |
| Pityriasis rosea | 347 | 278 | 69 | 69 |
| Psoriasis | 633 | 506 | 127 | 127 |
| Tinea Corporis | 385 | 308 | 77 | 77 |
| Varicella (chickepox) | 314 | 251 | 63 | 63 |
| Vitiligo | 323 | 258 | 65 | 65 |
| **Total** | **3406** | **2724** | **638** | **638** |

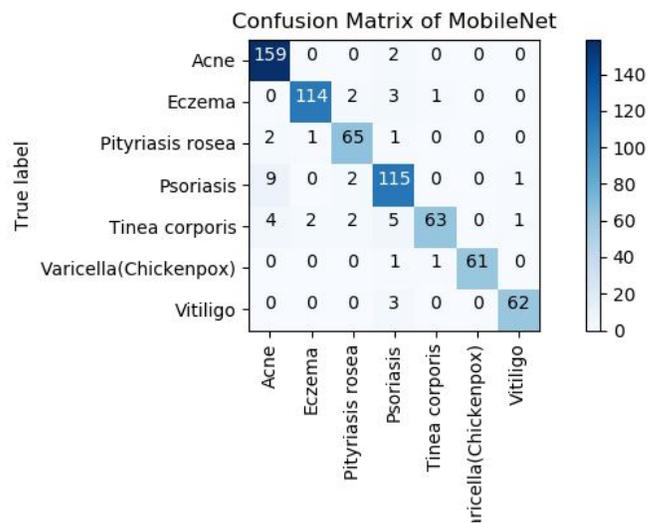

**Figure 5:** Confusion matrix using imbalanced dataset

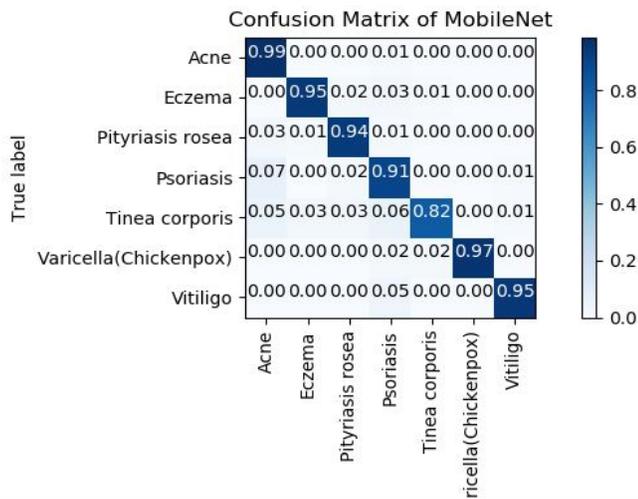

**Figure 6:** Normalized confusion matrix using imbalanced dataset

Figures 7-8 show the confusion matrix of the model without data augmentation and with oversampling technique used. This demonstrate how the system misclassify most of the psoriasis test images as acne however still the rank-1 accuracy of the model is 91.8%. Table 4 show how oversampling technique was used in the skin disease dataset.

**Table 4:** Dataset with oversampling technique

| Skin disease | Number of Images | Train Data (80%) | Test Data (20%) | Validation Data |
|---|---|---|---|---|
| Acne | 805 | 644 | 161 | 161 |
| Eczema | 599 | 644 | 161 | 161 |
| Pityriasis rosea | 347 | 644 | 161 | 161 |
| Psoriasis | 633 | 644 | 161 | 161 |
| Tinea Corporis | 385 | 644 | 161 | 161 |
| Varicella (chickepox) | 314 | 644 | 161 | 161 |
| Vitiligo | 323 | 644 | 161 | 161 |
| Total | 3406 | 4508 | 1127 | 1127 |

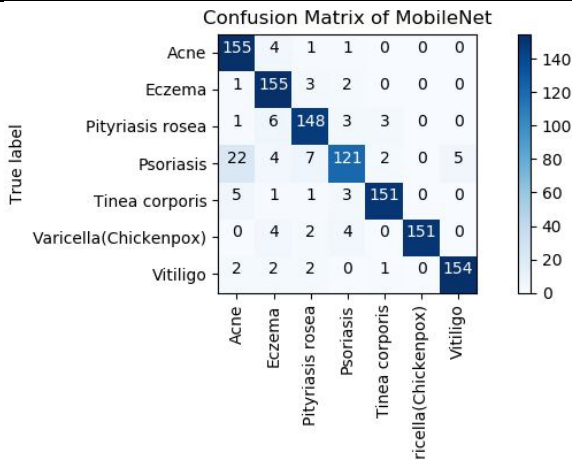

**Figure 7:** Confusion matrix without data augmentation

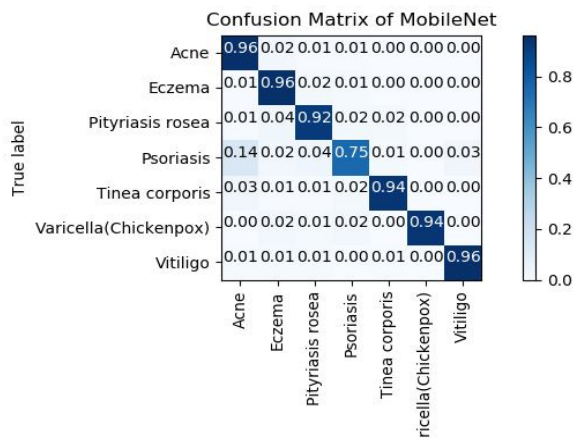

**Figure 8:** Normalized confusion matrix without data augmentation

Figures 9 to 10 show the confusion matrix of the model with data augmentation and with oversampling technique used. Still, this demonstrate how the system misclassify most of the psoriasis test images as acne and pityriasis rosea but from 75% accuracy on psoriasis, the accuracy goes up to 80% as well as the rank-1 accuracy of the model and is now, 94.4%. Table 5 shows how data augmentation is implemented on the skin disease dataset.

**Table 5:** Parameters used in data augmentation

| Process | Value |
|---|---|
| Rescale | 1./255 |
| Rotation range | 40 |
| Width shift range | 0.2 |
| Height shift range | 0.2 |
| Shear range | 0.2 |
| Zoom range | 0.2 |
| Horizontal flip | True |
| Fill mode | Nearest |

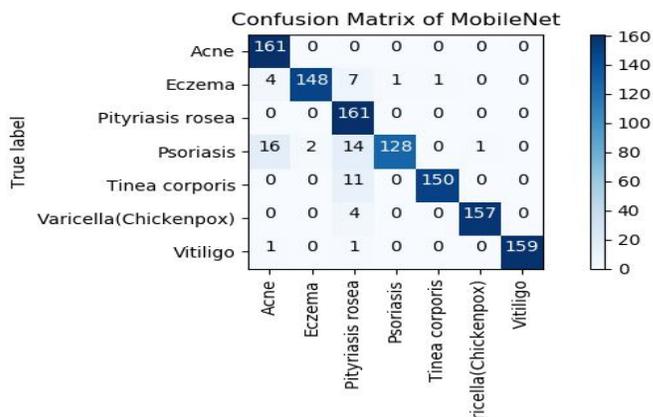

**Figure 9:** Confusion matrix with data augmentation

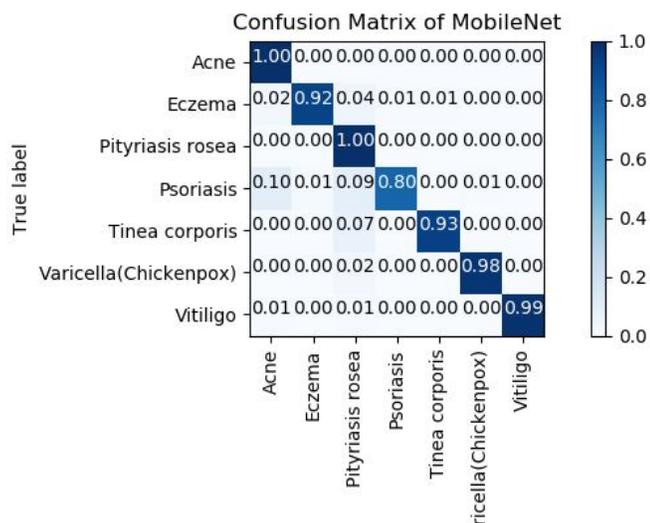

**Figure 10:** Normalized confusion matrix with data augmentation

### 3.2 Saliency Maps

Saliency maps can be used to visualize how the model predict each class with a given input [22]. The generated heatmap provide a way to intuitively visualize the location of the pixels where the model fixates most of its attention for diagnosis [23]. As can be seen, the model most set it focus where the lesion is located.

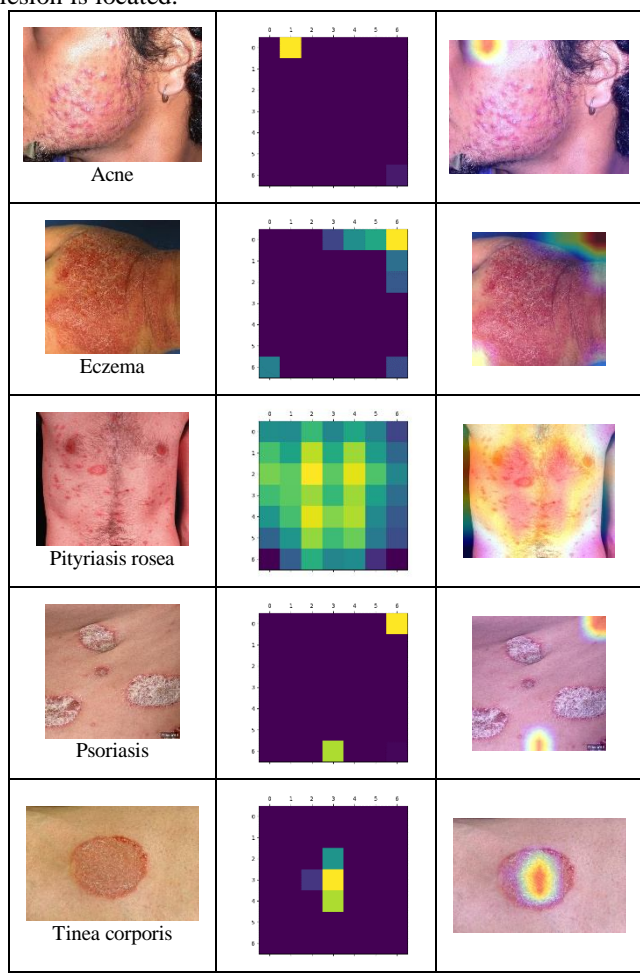

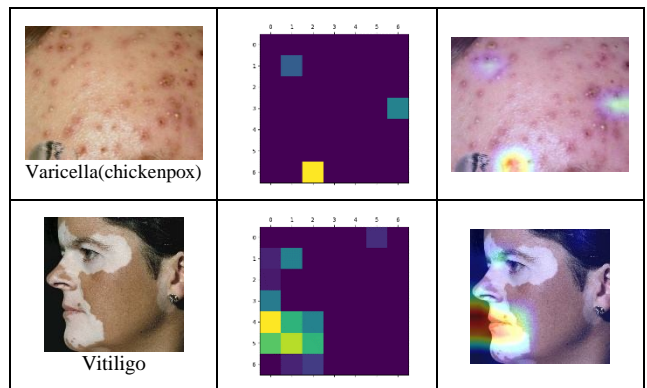

**Figure 11:** Saliency maps for 7 skin disease example images

## 4. CONCLUSION

The project achieved 94.4% accuracy in determining the seven skin diseases. Using undersampling method and the default preprocessing of input data achieved an 84.28% accuracy on the test dataset. While, using the imbalanced dataset and the default preprocessing of input data achieved a 93.6% accuracy. Then, the researcher used oversampling and the model attained a 91.8% accuracy. Lastly, using the oversampling and data augmentation technique provide an accuracy of 94.4%. In conclusion, in order to enhance the accuracy of the model different sampling techniques and preprocessing of input data can be explore. In our study, using oversampling and data augmentation generate the most accurate result.